\documentclass[letterpaper]{article}

\PassOptionsToPackage{square, numbers}{natbib}
\usepackage[final]{nips_2017}

\usepackage[utf8]{inputenc} 
\usepackage[T1]{fontenc}    
\usepackage{hyperref}       
\usepackage{url}            
\usepackage{booktabs}       
\usepackage{amsfonts}       
\usepackage{nicefrac}       
\usepackage{microtype}      

\usepackage{times}

\usepackage{wrapfig}

\usepackage{graphicx} 
\usepackage{caption}
\usepackage{subcaption}

\usepackage{algorithm}
\usepackage{algorithmic}

\usepackage{amsmath}
\usepackage{xfrac}

\usepackage{chngcntr}

\usepackage{multirow}
\usepackage{hhline}
\usepackage{siunitx}
\DeclareMathOperator*{\argmin}{argmin}

\DeclareMathOperator*{\minimize}{minimize}
\DeclareMathOperator*{\subjectto}{subject\;to}

\usepackage{color}

\newcommand\scalemath[2]{\scalebox{#1}{\mbox{\ensuremath{\displaystyle #2}}}}

\iftrue

\else

\fi

\title{Task-based End-to-end Model Learning\\in Stochastic Optimization} 

\author{
	Priya L. Donti \\
	Dept. of Computer Science \\
	Dept. of Engr. \& Public Policy \\
	Carnegie Mellon University \\
	Pittsburgh, PA 15213 \\
	\texttt{pdonti@cs.cmu.edu} \\
\And 
	Brandon Amos \\
	Dept. of Computer Science \\
	Carnegie Mellon University \\
	Pittsburgh, PA 15213 \\
	\texttt{bamos@cs.cmu.edu} \\
\And
	J. Zico Kolter \\
	Dept. of Computer Science \\
	Carnegie Mellon University \\
	Pittsburgh, PA 15213 \\
	\texttt{zkolter@cs.cmu.edu} \\
}

\begin{document}

\maketitle

\begin{abstract} 
With the increasing popularity of machine learning techniques,
it has become common to see prediction algorithms operating
within some larger process.
However, the criteria by which we train these algorithms often
differ from the ultimate criteria on which we evaluate them. 
This paper proposes an end-to-end approach for learning probabilistic
machine learning models in a manner that directly captures the ultimate task-based objective for which they 
will be used, 
within the context of stochastic programming.
We present three experimental
evaluations of the proposed approach: a classical inventory stock problem, 
a real-world electrical grid scheduling task,
and a real-world energy storage arbitrage task. 
We show that the proposed approach can outperform both
traditional modeling and purely black-box policy optimization approaches
in these applications.
\end{abstract} 

\section{Introduction} 
\label{sec:intro}
While prediction algorithms commonly operate
within some larger process,
the criteria by which we train these algorithms often
differ from the ultimate criteria on which we evaluate them: 
the performance of the full
``closed-loop'' system on the ultimate task at hand. 
For instance, instead of merely classifying images
in a standalone setting, one may want to use these classifications within 
planning and control tasks such as autonomous driving.  
While a typical image classification algorithm might optimize
accuracy or log likelihood, in a driving task we may ultimately 
care more about the difference between classifying a pedestrian as a tree
vs. classifying a garbage can as a tree. 
Similarly, when we use a probabilistic prediction algorithm to generate forecasts 
of upcoming electricity demand, we then want to use these forecasts 
to minimize the costs of a scheduling procedure that allocates  
generation for a power grid. 
As these examples suggest, 
instead of using a ``generic loss,'' we instead may want to learn
a model that approximates the ultimate task-based ``true loss.'' 

This paper considers an end-to-end approach for learning probabilistic
machine learning models that directly capture the objective
of their ultimate task.  
Formally, we consider
probabilistic models in the context of stochastic programming, 
where the goal is to minimize some expected cost over the models'
probabilistic predictions, subject to some (potentially also 
probabilistic) constraints.  As mentioned above, it is common to approach these
problems in a two-step fashion: first to fit a predictive model to observed data 
by minimizing some criterion such as negative log-likelihood,
and then to use this model to compute or approximate the necessary 
expected costs
in the stochastic programming setting.  While this procedure can work well in
many instances, it ignores the fact that the true cost of the system (the
optimization objective evaluated on \emph{actual} instantiations in the real
world) may benefit from a model that actually attains worse overall likelihood,
but makes more accurate predictions over certain manifolds of the
underlying space.

We propose to train a probabilistic model not (solely) for
predictive accuracy, but so that--when it is later used within the loop of a
stochastic programming procedure--it produces solutions that minimize the
ultimate task-based loss. 
This formulation may seem somewhat counterintuitive, given that a
``perfect'' predictive model would of course also be the optimal model to use
within a stochastic programming framework.
However, the reality that all models \emph{do} make errors illustrates 
that we should
indeed look to a final task-based objective to determine the proper 
error tradeoffs within a machine learning setting.
This paper proposes one way to evaluate task-based tradeoffs in
a fully automated fashion, by computing derivatives through the solution to the
stochastic programming problem in a manner that can improve the underlying
model.

We begin by presenting background material and related work in areas spanning
stochastic programming, end-to-end training, and optimizing alternative loss 
functions.  
We then describe our
approach within the formal context of stochastic programming, and give a
generic method for propagating task loss through these problems in
a manner that can update the models.  
We report on three experimental
evaluations of the proposed approach: a classical inventory 
stock problem, a real-world electrical grid scheduling task,
and a real-world energy storage arbitrage task.
We show that the proposed approach outperforms
traditional modeling and purely black-box policy optimization approaches.

\section{Background and related work}
\label{sec:background}
\paragraph{Stochastic programming}
Stochastic programming is a method for making decisions under uncertainty
by modeling or optimizing objectives governed by a random process.
It has applications in many domains such as energy 
\cite{wallace2003stochastic},
finance \cite{ziemba2006stochastic}, and manufacturing 
\cite{buzacott1993stochastic}, where the underlying
probability distributions are either known or can be estimated.
Common considerations 
include how to best model or approximate the underlying random variable, how to solve
the resulting optimization problem, and how to then assess the quality of the resulting (approximate) solution \cite{shapiro2007tutorial}.

In cases where the underlying probability distribution is known  but the
objective cannot be solved analytically, it is common to use Monte Carlo sample average
approximation methods, which draw multiple iid samples from  the underlying
probability distribution and then use deterministic optimization methods 
to solve the resultant problems
\cite{linderoth2006empirical}.  In cases where the underlying
distribution is not known, it is common to learn or estimate some model from
observed samples \cite{rockafellar1991scenarios}.

\paragraph{End-to-end training}
Recent years have seen a dramatic increase in the number of systems building on
so-called ``end-to-end'' learning.  Generally speaking, this term refers to
systems where the end goal of the machine learning process is directly predicted
from raw inputs \citep[e.g.][]{lecun2005off,thomas2006cognitive}.
In the context of deep learning systems, the term now
traditionally refers to architectures where, for example, there is no explicit
encoding of hand-tuned features on the data, but the system directly predicts
what the image, text, etc. is from the raw inputs
\cite{wang2011end,he2016deep,wang2012end,graves2014towards,amodei2015deep}.  
The context in which we use the term end-to-end is similar, but slightly more in
line with its older usage: instead of (just) attempting to learn an output (with
known and typically straightforward loss functions), we are specifically
attempting to learn a model based upon an end-to-end \emph{task} that the user is
ultimately trying to accomplish.  
We feel that this concept--of describing the entire closed-loop performance of the system as evaluated on the real task at hand--is beneficial to add to the notion of end-to-end learning.

Also highly related to our work are recent efforts in end-to-end policy learning
\cite{levine2016end}, using value iteration effectively as an
optimization procedure in similar networks \cite{tamar2016value}, and
multi-objective optimization \cite{harada2006local, van2014multi, mossalam2016multi, wiering2014model}.
These lines of work fit more with the ``pure'' end-to-end approach we discuss 
later on (where
models are eschewed for pure function approximation methods), but conceptually
the approaches have similar motivations in modifying typically-optimized policies to
address some task(s) directly.
Of course, the actual methodological
approaches are quite different, given our specific focus on stochastic
programming as the black box of interest in our setting. 

\paragraph{Optimizing alternative loss functions}
There has been a great deal of work in recent years on using machine learning
procedures to optimize different loss criteria
than those ``naturally'' optimized by the algorithm.  For example,
\citet{stoyanov2011empirical} and \citet{hazan2010direct} propose methods for
optimizing loss criteria in structured prediction that are \emph{different} from the
inference procedure of the prediction algorithm; 
this work has also recently been extended to deep networks \citep{song2016training}.
Recent work has also explored using auxiliary prediction losses to 
satisfy multiple objectives \cite{jaderberg2016reinforcement},
learning dynamics models that maximize control performance
in Bayesian optimization \cite{bansal2017goal},
and learning adaptive predictive models via differentiation through a meta-learning optimization objective \cite{finn2017model}.

The work we have found in the literature that most closely resembles our approach
 is the work of \citet{bengio1997using}, which uses
a neural network model for predicting financial prices, and then optimizes the
model based on returns obtained via a hedging strategy that employs it.  
We view this approach--of both using a model and then tuning that
model to adapt to a (differentiable) procedure--as a philosophical
predecessor to our own work.
In concurrent work, \citet{elmachtoub2017smart} also propose an approach for tuning model parameters given optimization results, but in the context of linear programming and outside the context of deep networks.
Whereas \citet{bengio1997using} and \citet{elmachtoub2017smart} 
use hand-crafted (but differentiable) algorithms to approximately
attain some objective given a predictive model, 
our approach is tightly coupled
to stochastic programming, where the explicit objective is to \emph{attempt} to
optimize the desired task cost via an exact optimization routine, but
given underlying randomness.
The notions of stochasticity are thus naturally quite
different in our work, but we do hope that our work can bring 
back the original idea of task-based model learning.
(Despite \citet{bengio1997using}'s original paper being nearly 20 years old,
virtually all follow-on work has focused on the financial application,
and not on what we feel is the core idea of using a surrogate model
within a task-driven optimization procedure.)

\section{End-to-end model learning in stochastic programming}
We first formally define the stochastic modeling and optimization problems
with which we are concerned.
Let ($x\in\mathcal{X},y\in\mathcal{Y}) \sim \mathcal
{D}$ denote standard input-output pairs drawn from some (real, unknown)
distribution $\mathcal{D}$.  
We also consider actions $z \in \mathcal{Z}$ that incur some
expected loss 
$L_\mathcal{D}(z) = E_{x,y \sim \mathcal{D}}[f(x,y,z)]$.
For instance, a power systems operator may try to allocate power
generators $z$ given past electricity demand $x$ 
and future electricity demand $y$;
this allocation's loss corresponds to the over- or under-generation penalties
incurred given future demand instantiations.

If we knew $\mathcal{D}$, then we could select optimal actions
$z_\mathcal{D}^\star = \argmin_z L_\mathcal{D}(z).$
However, in practice, the true distribution $\mathcal{D}$ is unknown.
In this paper, we are interested in modeling 
the conditional distribution $y|x$  using some parameterized model $p
(y|x;\theta)$ 
in order to minimize the real-world cost of the policy implied
by this parameterization.
Specifically, we find some parameters $\theta$ to parameterize
$p(y|x;\theta)$ (as in the standard statistical setting)
and then determine
optimal actions $z^\star(x;\theta)$ 
(via stochastic optimization)
that correspond to our observed input $x$ 
and the specific choice of parameters $\theta$ 
in our probabilistic model.
Upon observing the costs of these actions $z^\star(x;\theta)$ 
relative to true instantiations of $x$ and $y$, 
we update our parameterized model $p(y|x;\theta)$ accordingly, 
calculate the resultant new $z^\star(x;\theta)$, and repeat.
The goal is to find parameters $\theta$ such that the 
corresponding policy $z^\star (x;\theta)$ optimizes the loss
under the \emph{true} joint distribution of $x$ and $y$.

Explicitly, 
we wish to choose $\theta$ to minimize the \emph{task loss} $L(\theta)$
in the context of $x, y \sim \mathcal{D}$, i.e.
\begin{equation}
\label{eq:task-loss-unconst}
\minimize_\theta \;\; L(\theta) = \mathbf{E}_{x,y\sim\mathcal{D}} [ f(x,y,z^\star(x;\theta))].
\end{equation}
Since in reality we do not know the distribution $\mathcal{D}$,
we 
obtain $z^\star(x; \theta)$ via a proxy stochastic optimization problem
for a fixed instantiation of parameters $\theta$, i.e.
\begin{equation}
z^\star(x; \theta) = \argmin_z \;\; \mathbf{E}_{y \sim p(y | x;\theta)} [f(x, y, z)].
\label{eq:inner-opt-un}
\end{equation}
The above setting specifies $z^\star(x;\theta)$ using a simple (unconstrained) stochastic program, but in reality our decision may be subject to both probabilistic and deterministic constraints.  We therefore consider more general decisions produced through a generic stochastic programming problem\footnote{It is standard to presume in stochastic programming
	that equality constraints depend only on decision variables (not random variables), as non-trivial random equality constraints are typically
	not possible to satisfy.} 
\begin{equation}
\label{eq:inner-opt-const}
\begin{split}
z^\star(x;\theta) = \argmin_z \;\; &\mathbf{E}_{y \sim p(y | x;\theta)} [f(x, y, z)] \\
\subjectto \;\; &\mathbf{E}_{y \sim p(y | x;\theta)}[g_i(x, y,z)] \leq 0, \;\; i=1,\ldots,n_{ineq} \\
& h_i(z) = 0, \;\; i=1,\ldots,n_{eq}.
\end{split}
\end{equation}
In this setting, the full task loss is more complex, since it captures both the expected cost and any deviations from the constraints.  We can write this, for instance, as 
\begin{small}
	\begin{equation}
	\label{eq:task-loss-const}
	L(\theta) = \mathbf{E}_{x,y\sim\mathcal{D}} [ f(x,y,z^\star(x;\theta))]
	+ \sum_{i=1}^{n_{ineq}}I\{\mathbf{E}_{x,y\sim\mathcal{D}}[g_i(x,y,z^\star
	(x;\theta))] \leq 0\} 
	+\sum_{i=1}^{n_{eq}} \mathbf{E}_{x}[I\{h_i(z^\star(x;\theta)) = 0\}]
	\end{equation}
\end{small}%
(where $I(\cdot)$ is the indicator function that is zero when
its constraints are satisfied and infinite otherwise).  However, the basic intuition behind our approach remains the same for both the constrained and unconstrained cases: in both settings, we attempt to learn parameters of a probabilistic model not to produce strictly ``accurate'' predictions, but such that
\emph{when we use the resultant model within a stochastic programming setting, the resulting decisions perform well under the true distribution}.

Actually solving this problem requires that we differentiate through the ``argmin'' operator $z^\star
(x;\theta)$ of the stochastic programming problem.
This differentiation is not
possible for all classes of optimization problems (the argmin operator may be
discontinuous), but as we will show shortly, in many practical cases--including cases where the function
and constraints are strongly convex--we can indeed efficiently compute these gradients even in the context of constrained optimization.

\subsection{Discussion and alternative approaches}
We highlight our approach in contrast to two 
alternative
existing methods: traditional model learning and 
model-free black-box policy optimization. 
In traditional machine learning approaches, it is common to use $\theta$ to minimize the 
(conditional) log-likelihood of observed data under the model $p(y|x;\theta)$. 
This method corresponds to approximately solving the optimization problem
\begin{equation}
\minimize_\theta \;\; \mathbf{E}_{x,y \sim \mathcal{D}}\left[-\log p
(y|x;\theta)\right].
\label{eq:tradML}
\end{equation}
If we then need to use the conditional distribution $y|x$ to determine
actions $z$ within some later
optimization setting, we commonly use the predictive model obtained from \eqref{eq:tradML} directly.  
This
approach has obvious advantages, in that the model-learning phase is
well-justified independent of any future use in a task.  However, it is also
prone to poor performance in the common setting where the true
distribution $y|x$ cannot be represented within the class of distributions
parameterized by $\theta$, i.e. where the procedure suffers from model bias. 
Conceptually, the log-likelihood objective
\emph{implicitly} trades off between model error 
in different regions of the input/output space, but does so in a manner 
largely opaque to the modeler, and may ultimately \emph{not} 
employ the correct tradeoffs for a given task.

In contrast, there is an alternative approach to solving~\eqref{eq:task-loss-unconst} that we
describe as the model-free ``black-box'' policy optimization approach.  
Here, we forgo learning any model at all of the random variable $y$. 
Instead, we  attempt to learn a policy mapping directly from inputs
$x$ to actions $z^\star(x;\bar{\theta})$ that minimize 
the loss $L(\bar{\theta})$ presented in~\eqref{eq:task-loss-const}
(where here $\bar{\theta}$ defines the form of the policy itself, not
a predictive model).
While such model-free methods can perform well in many settings, 
they are often very data-inefficient, as the policy class must 
have enough representational power to describe sufficiently complex policies
without recourse to any underlying model.\footnote{This distinction is roughly
analogous to the policy search vs. model-based settings in reinforcement
learning. 
However, for the purposes of this paper, we
consider much simpler stochastic programs without the multiple rounds
that occur in RL, and the extension of these
techniques to a full RL setting remains as future work.}
\begin{wrapfigure}{l}{0.5\textwidth}
	\vspace{-23pt}
	\begin{minipage}{0.5\textwidth}
		\begin{algorithm}[H]
			\caption{Task Loss Optimization}
			\begin{algorithmic}[1]
				\STATE \textbf{input: } $\mathcal{D}$ \hspace{0.8em} \emph{// samples from true distribution}
				\STATE \textbf{initialize} $\theta$  \hspace{0.3em} \emph{// some initial parameterization}
				\vspace{5pt}
				\FOR{$t = 1, \ldots, T$}
				\STATE \textbf{sample} $(x, y) \sim \mathcal{D}$
				\STATE \textbf{compute} $z^\star(x;\theta)$ via Equation~\eqref{eq:inner-opt-const}
				\vspace{5pt}
				\STATE \emph{// step in violated constraint or objective}
				\IF{$\exists i$ s.t. $g_i(x,y,z^\star(x;\theta)) > 0$}
				\STATE \textbf{update}  $\theta$ with $\nabla_\theta g_i(x,y,z^\star
				(x;\theta))$
				\ELSE
				\STATE \textbf{update} $\theta$ with $\nabla_\theta f(x,y,z^\star
				(x;\theta))$
				\ENDIF
				\ENDFOR
			\end{algorithmic}
			\label{alg:task-loss-opt}
		\end{algorithm}
	\end{minipage}
	\vspace{-30pt}
\end{wrapfigure}

\vspace{-12pt}
Our approach offers an intermediate setting, where we \emph{do} still
use a surrogate model
to determine an optimal decision 
$z^\star(x; \theta)$, yet
we adapt this model 
based on the task loss instead
of any model prediction accuracy.  
In practice, we typically want to
minimize some weighted combination of log-likelihood \emph{and} task loss,
which can be easily accomplished given 
our approach.

\subsection{Optimizing task loss}
To solve the generic optimization problem \eqref{eq:task-loss-const}, we can in principle
adopt a straightforward (constrained) stochastic gradient approach, as detailed in
Algorithm~\ref{alg:task-loss-opt}. 
At each iteration, we solve the proxy stochastic programming problem \eqref{eq:inner-opt-const} to obtain $z^\star(x, \theta)$, using the distribution defined by 
our current values of
$\theta$.  
Then, we compute the true loss $L(\theta)$ using the
observed value of $y$. If any of the inequality constraints $g_i$ in $L(\theta)$
are violated, we take a gradient step in the violated constraint; otherwise, we take a gradient step in the optimization objective $f$.
We note that if any inequality constraints are probabilistic, Algorithm 1 must be adapted to employ mini-batches in order to determine whether these probabilistic constraints are satisfied.
Alternatively, because even the $g_i$ constraints are probabilistic, it is common in practice to simply move a weighted version of these constraints to the objective, i.e., we modify the objective by adding some 
 appropriate penalty times the positive part of the function, $\lambda g_i
(x,y,z)_+$, for some $\lambda > 0$.  
In practice, this has the effect of taking
gradient steps jointly in all the violated constraints and the objective in the
case that one or more inequality constraints are violated, often resulting in faster convergence.
Note that we need only move stochastic constraints into the objective; 
deterministic constraints on the policy itself will 
always be satisfied by the optimizer, as they are independent of the model.

\subsection{Differentiating the optimization solution to a stochastic
programming problem}

While the above presentation highlights the simplicity of the proposed approach,
it avoids the issue of chief technical challenge to this approach, which is computing
the gradient of an objective that depends upon the argmin operation $z^\star
(x;\theta)$.  Specifically, we need to compute the term
\begin{equation}
\frac{\partial L}{\partial \theta} = \frac{\partial L}{\partial z^\star} 
\frac{\partial z^\star}{\partial \theta} 
\end{equation}
which involves the Jacobian $\frac{\partial z^\star}{\partial \theta}$.  This is
the Jacobian of the optimal solution with respect to the distribution parameters
$\theta$.  Recent approaches have looked into similar argmin differentiations
\cite{gould2016differentiating, amos2016input},
though the methodology we present here is more general and handles the
stochasticity of the objective.

At a high level, we begin by writing the KKT optimality conditions of the general stochastic programming problem \eqref{eq:inner-opt-const}.
Differentiating these equations and applying the implicit function theorem gives
a set of linear equations that we can solve to obtain the necessary Jacobians
(with expectations over the 
distribution $y \sim p(y|x;\theta)$
denoted $\mathbf{E}_{y_\theta}$, and where $g$ is the vector of inequality constraints)
\begin{equation}
\begin{split}
&\left[
\renewcommand*{\arraystretch}{1.2}
\scalemath{0.75}{\begin{array}{ccc}
	\displaystyle \nabla_z^2 \mathbf{E}_{y_\theta} f(z) + \sum_{i=1}^{n_
		{ineq}}\lambda_i
	\nabla_z^2 \mathbf{E}_{y_\theta} g_i(z) & \left(\nabla_z \mathbf{E}_{y_\theta} g(z)\right)^T & A^T \\
	\mathrm{diag}(\lambda )\left(\nabla_z \mathbf{E}_{y_\theta} g
	(z)\right) & \mathrm
	{diag}(\mathbf{E}_{y_\theta} g(z)) & 0
	\\
	A & 0 & 0 \end{array}} \right ]
\renewcommand*{\arraystretch}{1.5}
\left [ \scalemath{0.85}{\begin{array}{c} 
	\frac{\partial z}{\partial \theta} \\
	\frac{\partial \lambda}{\partial \theta} \\
	\frac{\partial \nu}{\partial \theta}
	\end{array}} \right ]
\renewcommand*{\arraystretch}{1.5}
= -\left [ \scalemath{0.85}{\begin{array}{c}
	\frac{\partial \nabla_z \mathbf{E}_{y_\theta} f(z)}{\partial \theta} +
	\frac { \partial \sum_{i=1}^{n_{ineq}} \lambda_i \nabla_z \mathbf
		{E}_{y_\theta} g_i(z)}
	{\partial \theta} \\
	\mathrm{diag}(\lambda) \frac{\partial \mathbf{E}_{y_\theta} g(z)}{\partial \theta} \\
	0
	\end{array}} \right ].
\end{split}
\label{eq-jacobian-equation}
\end{equation}
The terms in these equations look somewhat complex, but fundamentally, the left side gives
the optimality conditions of the convex problem, and
the right side gives the derivatives of the relevant functions
at the achieved solution with respect to the governing parameter $\theta$.   
In practice, we calculate the right-hand terms by employing sequential quadratic
programming \cite{boggs1995sequential} to find the optimal policy $z^\star(x;\theta)$
for the given parameters $\theta$, using a 
recently-proposed approach for fast solution of the argmin
differentiation for QPs \cite{amos2017optnet} to solve the necessary 
linear equations; we then take the derivatives at the optimum produced by this strategy.
Details of this approach are described in the appendix.

\section{Experiments}
\label{sec:experiments}

\begin{figure*}[t!]
	\centering
	\begin{subfigure}[t]{0.32\textwidth}
		\centering
		\includegraphics[width=\textwidth]{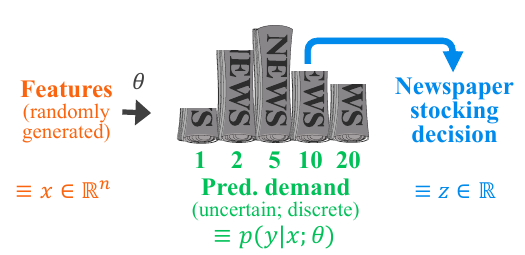}
		\caption{Inventory stock problem}
		\label{fig:newsvendor-intro}
	\end{subfigure}%
	~ 
	\begin{subfigure}[t]{0.32\textwidth}
		\centering
		\includegraphics[width=\textwidth]{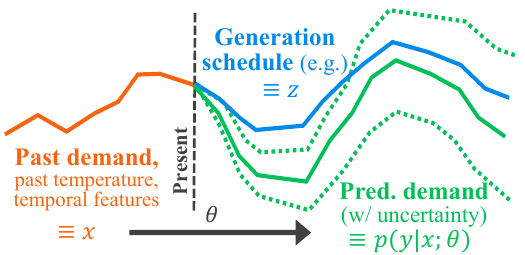}
		\caption{Load forecasting problem}
		\label{fig:power-sched-intro}
	\end{subfigure}
	~
	\begin{subfigure}[t]{0.32\textwidth}
			\centering
			\includegraphics[width=\textwidth]{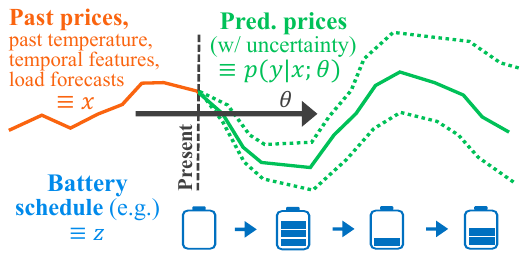}
			\caption{Price forecasting problem}
			\label{fig:storage-intro}
	\end{subfigure}
	\caption{Features $x$, model predictions $y$, and policy $z$ for the three experiments.}
	\label{fig:variables-intro}
	\vspace{-0.5em}
\end{figure*}

We consider three applications of our task-based method: a
synthetic inventory stock problem, a real-world energy scheduling task, and a real-world battery arbitrage task. 
We demonstrate that
the task-based end-to-end approach can substantially improve upon other
alternatives.
Source code for all experiments is available at \url{https://github.com/locuslab/e2e-model-learning}.

\subsection{Inventory stock problem}

\paragraph{Problem definition} To highlight the performance of the
algorithm in a setting where
the true underlying model is known to us, we consider a ``conditional''
variation of the classical inventory stock problem \cite{shapiro2007tutorial}.  
In this problem, 
a company must order some quantity $z$ of a product to 
minimize costs over some stochastic demand $y$, whose distribution in turn is
affected by some observed features $x$ (Figure~\ref{fig:newsvendor-intro}).  There are linear and quadratic costs on the amount of
product ordered, plus different linear/quadratic costs on over-orders $[z-y]_+$
and under-orders $[y-z]_+$.  The objective is given by
\begin{equation}
f_{stock}(y, z) = c_0z + \frac{1}{2}q_0 z^2 + c_b[y-z]_{+} 
+  \frac{1}{2}q_b ([y-z]_{+})^2 
+ c_h[z-y]_{+} +  \frac{1}{2} q_h ([z-y]_{+})^2,
\end{equation}
where $[v]_+ \equiv \max\{v, 0\}$. 
For a specific choice of probability model $p(y|x;\theta)$, 
our proxy stochastic programming problem can then be written as
\begin{equation}
\minimize_z \;\; \mathbf{E}_{y \sim p(y | x;\theta)} [f_{stock}(y,z)].
\label{eq-obj-inv}
\end{equation}
To simplify the setting, we further assume that the demands are discrete, taking
on values $d_1,\ldots,d_k$ with probabilities (conditional on $x$) $(p_\theta)_i
\equiv p (y = d_i | x;\theta)$.  Thus our stochastic programming problem \eqref{eq-obj-inv} 
can be written succinctly as a joint
quadratic program\footnote{This is referred to as a two-stage stochastic
programming problem (though a very trivial example of one), where 
first stage variables consist of the amount of product to buy before observing
demand, and second-stage variables consist of how much to sell back or
additionally purchase once the true demand has been revealed.}
\begin{equation}
\begin{split}
\minimize_{z \in \mathbb{R}, z_b, z_h \in \mathbb{R}^k} \;\; & 
c_0 z + \frac{1}{2}q_0 z^2 + 
\sum_{i=1}^k (p_\theta)_i \left( c_b (z_b)_i + \frac{1}{2}q_b 
(z_b)_i^2
+ c_h (z_h)_i + \frac{1}{2}q_h (z_h)_i^2
\right ) \\
\subjectto \;\; & d - z\boldsymbol{1} \leq z_b,  \;\; z\boldsymbol{1} - d \leq z_h, \;\;
z,z_h, z_b \geq 0.
\end{split}
\end{equation}
Further details of this approach are given in the appendix.

\paragraph{Experimental setup} 
We examine our algorithm under two main conditions: 
where the true model is linear, and where it is nonlinear.
In all cases, we generate problem instances by randomly sampling some $x \in
\mathbb{R}^n$ and then generating $p(y|x;\theta)$ according to either
$p(y|x;\theta) \propto \exp(\Theta^Tx)$ (linear true model) or $p(y|x;\theta) \propto \exp(
(\Theta^Tx)^2)$  (nonlinear true model) 
for some $\Theta \in \mathbb{R}^{n\times k}$.  
We compare the following approaches on these tasks: 
1) the QP allocation based upon the true
model (which performs optimally); 
2) MLE approaches (with
linear or nonlinear probability models) that fit a model to the
data, and then compute the allocation by solving the QP;
3) pure end-to-end policy-optimizing models 
(using linear or nonlinear hypotheses for the policy); 
and 
4) our task-based learning models (with linear or nonlinear
probability models).
\begin{figure*}[t!]
	\vskip 0.2in
	\centering
	\includegraphics[width=0.95\textwidth]{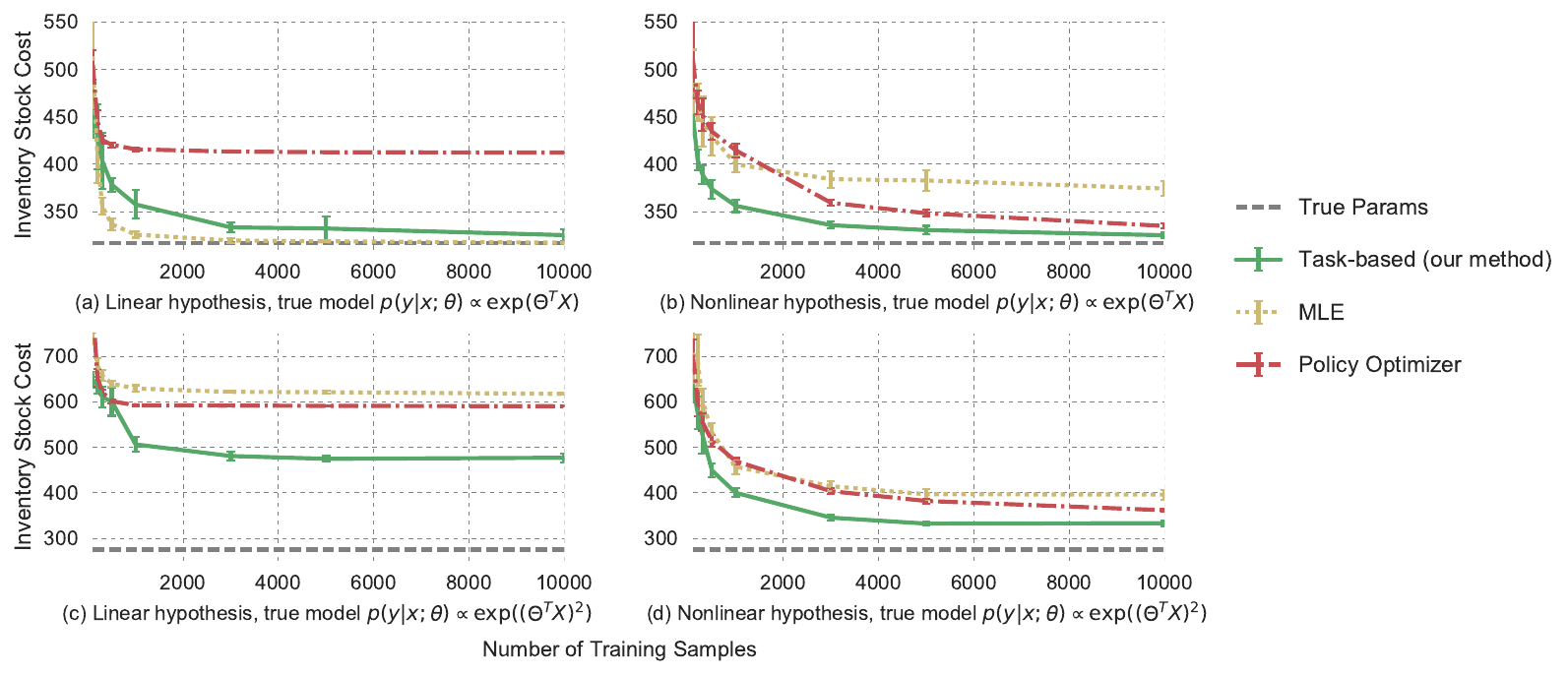}
	\caption{Inventory problem results for 10 runs over a representative instantiation of true parameters ($c_0 = 10, q_0 = 2, c_b = 30, q_b = 14, c_h = 10, q_h=2$). 
	Cost is evaluated over 1000 testing samples (lower is better).
		The linear MLE performs best for a true linear model.
		In all other cases, the task-based models outperform their 
		MLE and policy counterparts.}
	\vskip -0.2in
	\label{fig:inventory-results}
\end{figure*} 
In all cases, we evaluate test performance by
running on 1000 random examples, and evaluate performance over 10 folds of
different true $\theta^\star$ parameters.

Figures~\ref{fig:inventory-results}(a) and (b) show the performance of these
methods given a linear true model, 
with linear and nonlinear model hypotheses, respectively.
As expected, the linear MLE approach performs best, as the true underlying
model is in the class of distributions that it can represent and thus
solving the stochastic programming problem is a very strong proxy for solving
the true optimization problem under the real distribution.
While the true model is also contained within the nonlinear MLE's 
generic nonlinear distribution class, we see that this method requires more
data to converge, and when given less data makes 
error tradeoffs that are ultimately not the
correct tradeoffs for the task at hand; our task-based approach
thus outperforms this approach.
The task-based approach also substantially outperforms 
the policy-optimizing neural network,
highlighting the fact that it is more data-efficient to run the learning process
``through'' a reasonable model.  Note that here it does not make a difference
whether we use the linear or nonlinear model in the task-based approach.

Figures~\ref{fig:inventory-results}(c) and (d) show performance 
in the case of a nonlinear true model, 
with linear and nonlinear model hypotheses, respectively.
Case (c) represents the ``non-realizable'' case, where the true underlying distribution
cannot be represented by
the model hypothesis class.
Here, the linear MLE, as expected, performs very poorly: it cannot capture the true
underlying distribution, and thus the resultant stochastic programming solution
would not be expected to perform well.  
The linear policy model similarly performs poorly.
Importantly, the task-based approach
with the \emph{linear} model performs much better here: despite the fact that it
still has a misspecified model, the task-based nature of the learning process
lets us learn a \emph{different} linear model than the MLE version, which is
particularly tuned to the distribution and loss of the task.  
Finally, also as to be expected, the non-linear models perform better 
than the linear models in this scenario, 
but again with the task-based non-linear model outperforming the nonlinear MLE and 
end-to-end policy approaches.

\subsection{Load forecasting and generator scheduling}
\label{subsec:load}
We next consider a more realistic grid-scheduling task, based upon over 8 years
of real electrical grid data. In this setting, a power system operator must 
decide how much electricity generation $z \in \mathbb{R}^{24}$ to schedule for 
each hour in the next 24 hours based on some 
(unknown) distribution over electricity demand (Figure~\ref{fig:power-sched-intro}).  
Given a particular realization $y$ of demand, 
we impose penalties for both generation excess ($\gamma_e$)
and generation shortage ($\gamma_s$),
with $\gamma_s \gg \gamma_e$.
We also add a quadratic regularization term, indicating a preference for
generation schedules that closely match demand realizations.
Finally, we impose a ramping constraint $c_r$ restricting 
the change in generation between consecutive timepoints,
reflecting physical limitations associated with quick changes in
electricity output levels.  These are reasonable proxies for the actual economic
costs incurred by electrical grid operators when scheduling generation, and
can be written as the stochastic programming problem
\begin{equation}
\begin{split}
\minimize_{z \in \mathbb{R}^{24}} \;\;& \sum_{i=1}^{24} \mathbf{E}_{y \sim p(y | x;\theta)}\left [ \gamma_s [y_i-z_i]_+ + \gamma_e
[z_i-y_i]_+ + \frac{1}{2}(z_i - y_i)^2 \right ]\\
\subjectto \;\;& |z_i - z_{i-1}| \leq c_r \; \forall i,
\end{split}
\label{eq-gen-optim-body}
\end{equation}
where $[v]_+ \equiv \max\{v, 0\}$. Assuming (as we will in our model), that $y_i$ is a Gaussian random variable with mean $\mu_i$
and variance $\sigma_i^2$, then this expectation has a closed form that can be
computed via analytically integrating the Gaussian PDF.\footnote{
	Part of the philosophy behind applying this approach here is that we
	\emph{know} the Gaussian assumption 
	is incorrect: the true underlying load 
	is neither Gaussian distributed nor homoskedastic.
	However, these
	assumptions are exceedingly common in practice, 
	as they enable easy model learning and exact analytical solutions.
	Thus, training the
	(still Gaussian) system with a task-based loss 
	retains computational tractability while still allowing us to
	modify the distribution's parameters to improve actual performance
	on the task at hand.
	}
We then use sequential quadratic programming (SQP) to iteratively
approximate the resultant convex objective as a quadratic objective, iterate until
convergence, and then compute the necessary Jacobians using the quadratic
approximation at the solution, which gives the correct Hessian and gradient
terms. 
Details are given in the appendix.

\begingroup
	\centering
\setcounter{figure}{3}
\begin{figure*}[t!]
	\centering
	\vskip 0.2in
	\includegraphics[width=0.95\textwidth]{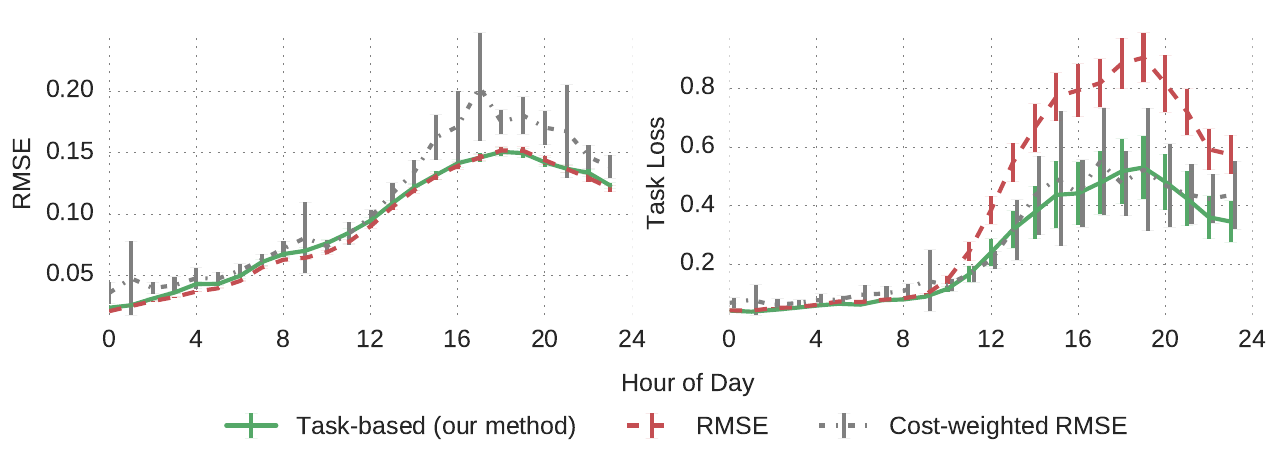}
	\caption{Results for 10 runs of the generation-scheduling problem for representative decision parameters $\gamma_e = 0.5, \gamma_s = 50,$ and $c_r = 0.4$.
		(Lower loss is better.)
		As expected, the RMSE net achieves the lowest RMSE for its predictions.
		However, the task net outperforms the RMSE net on task loss by 38.6\%, 
		and the cost-weighted RMSE on task loss by 8.6\%.}
	\vskip -0.2in
	\label{fig-forecasting}
\end{figure*} 
\endgroup

To develop a predictive model, we make use of
a highly-tuned load forecasting methodology. 
Specifically, we input 
the past day's electrical load and
temperature, the next day's temperature forecast, and
additional features such as non-linear functions of the temperatures, binary
indicators of weekends or holidays, and yearly sinusoidal features. We then predict
the electrical load over all 24 hours of the next day.  
We employ a 2-hidden-layer
neural network for this purpose,
with an additional residual connection from
the inputs to the outputs initialized to the linear regression solution. 
\begin{wrapfigure}{l}{0.45\textwidth}
	\begingroup
	\setcounter{figure}{2}  
	\vspace{-18pt}
	\begin{minipage}{0.43\textwidth}
		\begin{figure}[H]
			\vskip 0.2in
			\includegraphics[width=\textwidth]{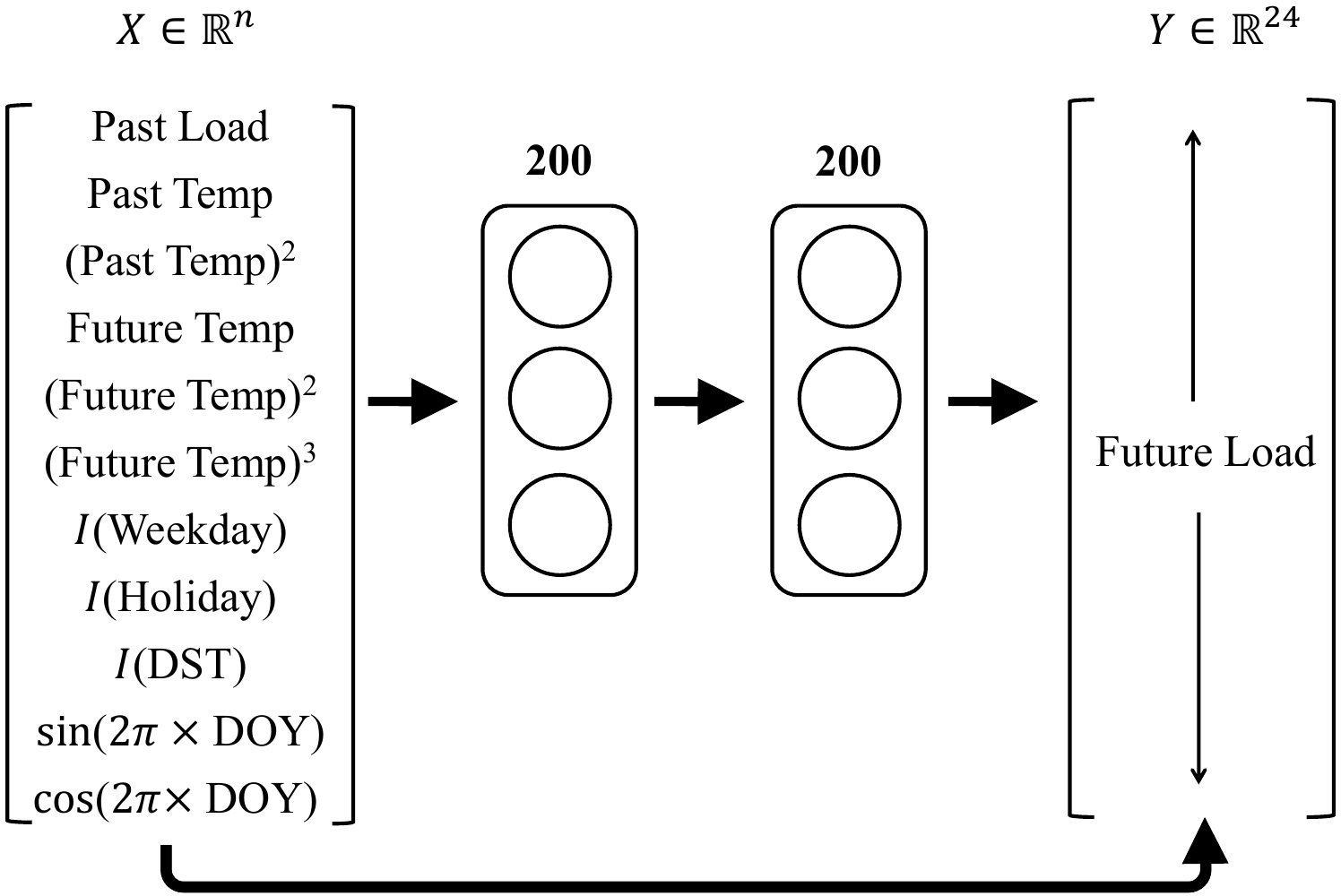}
			\caption{2-hidden-layer neural network to predict hourly electric load for the next day.}
			\label{fig-arch}
		\end{figure} 
	\end{minipage}
	\vspace{-10pt}
	\endgroup
\end{wrapfigure}
An illustration of the architecture is shown
in Figure~\ref{fig-arch}.
We train the model to minimize the mean squared
error between its predictions and the actual load (giving the mean
prediction $\mu_i$), and compute $\sigma^2_i$ as the (constant) empirical 
variance between the predicted and actual values.
In all cases we use 7 years of data to train the model, and 1.75 subsequent years for testing.

Using the (mean and variance) predictions of this base model, we obtain $z^\star(x;\theta)$ by solving the generator scheduling problem \eqref{eq-gen-optim-body} and then adjusting network parameters to minimize the resultant task loss.
We compare against a traditional stochastic programming model that minimizes just the RMSE, as well as a cost-weighted RMSE that periodically reweights training samples given their task loss.\footnote{It is worth noting that a cost-weighted RMSE approach is only possible when direct costs can be assigned independently to each decision point, i.e. when costs do not depend on multiple decision points (as in this experiment). Our task-based method, however, accommodates the (typical) more general setting.}
 (A pure policy-optimizing network is not shown, as it could not sufficiently learn the ramp constraints. 
We could not obtain good performance for the policy optimizer even ignoring this infeasibility.)

Figure~\ref {fig-forecasting} shows the performance of the three models on the testing dataset.
As expected, the RMSE model
performs best with respect to the RMSE of its predictions
(its objective).
However, the task-based model substantially
outperforms the RMSE model when evaluated on task loss, the actual
objective that the system operator cares about:
specifically, we improve upon the performance of the
traditional stochastic programming method by 38.6\%.
The cost-weighted RMSE's performance is extremely variable, and overall,
the task net improves upon this method by 8.6\%.

\subsection{Price forecasting and battery storage}
\label{subsec:storage}

Finally, we consider a battery arbitrage task, based upon 6 years of real electrical grid data.
Here, a grid-scale battery must operate over a 24 hour period based on some (unknown) distribution over future electricity prices (Figure~\ref{fig:storage-intro}).
For each hour, the operator must decide how much to charge ($z_{\text{in}} \in \mathbb{R}^{24}$) or discharge ($z_{\text{out}} \in \mathbb{R}^{24}$) the battery, thus inducing a particular state of charge in the battery ($z_{\text{state}} \in \mathbb{R}^{24}$).
Given a particular realization $y$ of prices, the operator optimizes over: 1) profits, 2) flexibility to participate in other markets, by keeping the battery near half its capacity $B$ (with weight $\lambda$), and 3) battery health, by discouraging rapid charging/discharging (with weight $\epsilon$, $\epsilon < \lambda$).
The battery also has a charging efficiency ($\gamma_{\text{eff}}$), limits on speed of charge ($c_{\text{in}}$) and discharge ($c_{\text{out}}$), and begins at half charge.
This can be written as the stochastic programming problem
\begin{equation}
\begin{split}
\minimize_{z_{\text{in}}, z_{\text{out}}, z_{\text{state}} \in \mathbb{R}^{24}} \;\;& \mathbf{E}_{y \sim p(y | x;\theta)}\left [ \sum_{i=1}^{24} y_i (z_{\text{in}} - z_{\text{out}})_i
+ \lambda \left\lVert z_{\text{state}} - \frac{B}{2} \right\rVert^2
+ \epsilon \lVert z_{\text{in}} \rVert^2 + \epsilon \lVert z_{\text{out}} \rVert^2
\right ]\\
\subjectto \;\;& z_{\text{state}, i+1} = z_{\text{state}, i} -  z_{\text{out}, i} + \gamma_{\text{eff}}z_{\text{in}, i} \; \forall i, \;\; z_{\text{state}, 1} = B/2, \\
& 0 \leq z_{\text{in}} \leq c_{\text{in}}, \;\; 0 \leq z_{\text{out}} \leq c_{\text{out}}, \;\; 0 \leq z_{\text{state}} \leq B.
\end{split}
\label{eq-storage-optim-body}
\end{equation}
Assuming (as we will in our model) that $y_i$ is a random variable with mean $\mu_i$, then this expectation has a closed form that depends only on the mean.
Further details are given in the appendix.

To develop a predictive model for the mean, we use an architecture similar to that described in Section~\ref{subsec:load}.
In this case, we input the past day's prices and temperature, the next day's load forecasts and temperature forecasts, and additional features such as non-linear functions of the temperatures and temporal features similar to those in Section~\ref{subsec:load}.
We again train the model to minimize the mean squared error between the model's predictions and the actual prices (giving the mean prediction $\mu_i$), using about 5 years of data to train the model and 1 subsequent year for testing.
Using the mean predictions of this base model, we then solve the storage scheduling problem by solving the optimization problem (\ref{eq-storage-optim-body}), again learning network parameters by minimizing the task loss.
We compare against a traditional stochastic programming model that minimizes just the RMSE.

\begin{table}
	\centering
\setlength{\tabcolsep}{0.5em} 
{\renewcommand{\arraystretch}{1.2}
	{\small
	\begin{tabular}{| c | c | S[table-number-alignment=center, table-figures-uncertainty=3, round-mode=figures, round-precision=3] | S[table-number-alignment=center, table-figures-uncertainty=3, round-mode=figures, round-precision=3] | c | }
		\hline
		\multicolumn{2}{|c|}{\textbf{Hyperparameters}} & \multicolumn{1}{c|}{\multirow{2}{*}{\textbf{RMSE net}}} & \multicolumn{1}{c|}{\multirow{2}{*}{\textbf{Task-based net (our method)}}} & \multirow{2}{*}{\textbf{\% Improvement}} \\
		\hhline{--~~~}
		$\boldsymbol{\lambda}$ & $\boldsymbol{\epsilon}$ &  & & \\ 
		\hline
		0.1 & 0.05 & -1.45 \pm 4.67 & -2.92 \pm 0.3 & 102  \\ \hline
		1 & 0.5 & 4.96 \pm 4.85 & 2.28 \pm 2.99 & 54 \\ \hline
		10 & 5 & {$\;\;\;\:131 \pm 145$} & {$\;\;\;\,95.9 \pm 29.8$} & 27 \\ \hline
		35 & 15 & {$\;\;\;\;\: 173 \pm 7.38$} & {$\;\;\;\;\,170 \pm 2.16$} & 2 \\ \hline
	\end{tabular}}}
\vspace{1em}
\caption{Task loss results for 10 runs each of the battery storage problem, given a lithium-ion battery with attributes $B=1$, $\gamma_{\text{eff}} = 0.9$, $c_{\text{in}} = 0.5$, and $c_{\text{out}} = 0.2$. (Lower loss is better.) Our task-based net on average somewhat improves upon the RMSE net, and demonstrates more reliable performance.}
\vspace{-0.5em}
\label{table:storage-results}
\end{table}

Table~\ref{table:storage-results} shows the performance of the two models on the testing dataset.
As energy prices are difficult to predict due to numerous outliers and price spikes, the models in this case are not as well-tuned as in our load forecasting experiment; thus, their performance is relatively variable.
Even then, in all cases, our task-based model demonstrates better average performance than the RMSE model when evaluated on task loss, the objective most important to the battery operator (although the improvements are not statistically significant).
More interestingly, our task-based method shows less (and in some cases, far less) variability in performance than the RMSE-minimizing method.
Qualitatively, our task-based method hedges against perverse events such as price spikes that could substantially affect the performance of a battery charging schedule.
The task-based method thus yields more reliable performance than a pure RMSE-minimizing method in the case the models are inaccurate due to a high level of stochasticity in the prediction task.

\section{Conclusions and future work}
\label{sec:conclusion}
This paper proposes an end-to-end approach for learning
machine learning models that will be used in the loop of
a larger process. 
Specifically, we consider training probabilistic models 
in the context of stochastic programming to directly capture 
a task-based objective.
Preliminary experiments indicate that our task-based learning model
substantially outperforms MLE and policy-optimizing approaches in all 
but the (rare) case that the
MLE model ``perfectly'' characterizes the underlying distribution.
Our method also achieves a 38.6\% performance improvement over a
highly-optimized real-world stochastic programming algorithm 
for scheduling electricity generation based on predicted load.
In the case of energy price prediction, 
where there is a high degree of inherent stochasticity in the problem,
our method demonstrates more reliable task performance than a traditional predictive method.
The task-based approach thus demonstrates promise in 
optimizing in-the-loop predictions.
Future work includes an extension of our approach to stochastic learning models
with multiple rounds, and further to model predictive control and full reinforcement learning settings.

\newpage

\section*{Acknowledgments}
This material is based upon work supported by the National Science Foundation Graduate Research Fellowship Program under Grant No. DGE1252522, and by the Department of Energy Computational Science Graduate Fellowship under Grant No. DE-FG02-97ER25308. We thank Arunesh Sinha for providing helpful corrections.

\bibliographystyle{unsrtnat}
\bibliography{references}

\newpage
\clearpage
\pagenumbering{arabic}
\renewcommand*{\thepage}{A\arabic{page}}
\appendix
\counterwithin{figure}{section}
\counterwithin{table}{section}
\counterwithin{equation}{section}
\counterwithin{footnote}{section}

\section{Appendix}
\label{sec:appendix}
We present some computational and architectural
details for the proposed 
task-based learning model, both in the general case and for the experiments described in Section~\ref{sec:experiments}.

\subsection{Differentiating the optimization solution to a stochastic programming problem}

The issue of chief technical challenge to our approach is computing
the gradient of an objective that depends upon the argmin operation $z^\star
(x;\theta)$.  Specifically, we need to compute the term
\begin{equation}
\frac{\partial L}{\partial \theta} = \frac{\partial L}{\partial z^\star} 
\frac{\partial z^\star}{\partial \theta} 
\end{equation}
which involves the Jacobian $\frac{\partial z^\star}{\partial \theta}$.  This is
the Jacobian of the optimal solution with respect to the distribution parameters
$\theta$.  Recent approaches have looked into similar argmin differentiations
\cite{gould2016differentiating, amos2016input},
though the methodology we present here is more general and handles the
stochasticity of the objective.

We begin by writing the KKT optimality conditions of the general stochastic
programming problem \eqref{eq:inner-opt-const}, where all expectations are taken with
respect to the modeled distribution $y \sim p(y|x;\theta)$ (for compactness, denoted here as $\mathbf{E}_{y_\theta}$).
Further, assuming the problem is convex means we can
replace the general equality constraints $h(z) = 0$ with the linear constraint
$Az = b$.  A point $(z,\lambda,\nu)$ is a primal-dual optimal point
if it satisfies
\begin{equation}
\begin{split}
\mathbf{E}_{y_\theta} g(z) & \leq 0 \\
Az & = b \\
\lambda & \geq 0 \\
\lambda \circ \mathbf{E}_{y_\theta} g(z) & = 0 \\
\nabla_z \mathbf{E}_{y_\theta} f(z) + \lambda^T \nabla_z \mathbf{E}_{y_\theta} g
(z) + A^T \nu & = 0
\end{split}
\end{equation}
where here $g$ denotes the vector of all inequality constraints (represented as a
vector-valued function), and where we wrap the dependence on $x$ and $y$ into the
functions $f$ and $g_i$ themselves.

Differentiating these equations and applying the implicit function theorem gives
a set of linear equations that we can solve to obtain the necessary Jacobians
\begin{equation}
\begin{split}
&\left[
\renewcommand*{\arraystretch}{1.2}
\scalemath{0.75}{\begin{array}{ccc}
	\displaystyle \nabla_z^2 \mathbf{E}_{y_\theta} f(z) + \sum_{i=1}^{n_
		{ineq}}\lambda_i
	\nabla_z^2 \mathbf{E}_{y_\theta} g_i(z) & \left(\nabla_z \mathbf{E}_{y_\theta} g(z)\right)^T & A^T \\
	\mathrm{diag}(\lambda )\left(\nabla_z \mathbf{E}_{y_\theta} g
	(z)\right) & \mathrm
	{diag}(\mathbf{E}_{y_\theta} g(z)) & 0
	\\
	A & 0 & 0 \end{array}} \right ]
\renewcommand*{\arraystretch}{1.5}
\left [ \scalemath{0.85}{\begin{array}{c} 
	\frac{\partial z}{\partial \theta} \\
	\frac{\partial \lambda}{\partial \theta} \\
	\frac{\partial \nu}{\partial \theta}
	\end{array}} \right ]
\renewcommand*{\arraystretch}{1.5}
= -\left [ \scalemath{0.85}{\begin{array}{c}
	\frac{\partial \nabla_z \mathbf{E}_{y_\theta} f(z)}{\partial \theta} +
	\frac { \partial \sum_{i=1}^{n_{ineq}} \lambda_i \nabla_z \mathbf
		{E}_{y_\theta} g_i(z)}
	{\partial \theta} \\
	\mathrm{diag}(\lambda) \frac{\partial \mathbf{E}_{y_\theta} g(z)}{\partial \theta} \\
	0
	\end{array}} \right ].
\end{split}
\label{eq-jacobian-equation-app}
\end{equation}
The terms on the left side are the optimality conditions of the convex problem, and
the terms on right side are the derivatives of the relevant functions
at the achieved solution, with respect to the governing parameter $\theta$.  
These equations will take slightly different forms depending on how the stochastic
programming problem is solved, but are usually fairly straightforward to
compute if the solution is solved in some ``exact'' manner (i.e., where second
order information is used).  
In practice, we calculate the right side of this equation by employing sequential quadratic
programming \cite{boggs1995sequential} to find the optimal policy $z^\star$
for the given parameters $\theta$, using a 
recently-proposed approach for fast solution of argmin
differentiation for QPs \cite{amos2017optnet} to solve the necessary 
linear equations; we then take the derivatives at the optimum produced by this strategy.

\subsection{Details on computation for inventory stock problem}
The objective for our ``conditional'' variation of the classical 
inventory stock problem is
\begin{equation}
f_{stock}(y, z) = c_0z + \frac{1}{2}q_0 z^2 + c_b[y-z]_{+} 
+  \frac{1}{2}q_b ([y-z]_{+})^2 
+ c_h[z-y]_{+} +  \frac{1}{2} q_h ([z-y]_{+})^2
\end{equation}
where $z$ is the amount of product ordered; $y$ is the stochastic electricity demand (which is affected by features $x$); $[v]_+ \equiv \max\{v, 0\}$; and $(c_0, q_0), (c_b, q_b),$ and $(c_h,q_h)$ are linear and quadratic costs on the amount of product ordered, over-orders, and under-orders, respectively.
Our proxy stochastic programming problem can then be written as
\begin{equation}
\minimize_z \;\; L(\theta) = \mathbf{E}_{y \sim p(y | x;\theta)} [f_{stock}(y,z)].
\label{eq-obj-inv-app}
\end{equation}
To simplify the setting, we further assume that the demands are discrete, taking
on values $d_1,\ldots,d_k$ with probabilities (conditional on $x$) $(p_\theta)_i
\equiv p (y = d_i | x;\theta)$.  Thus our stochastic programming problem \eqref{eq-obj-inv-app} 
can be written succinctly as a joint
quadratic program
\begin{equation}
\begin{split}
\minimize_{z \in \mathbb{R}, z_b, z_h \in \mathbb{R}^k} \;\; & 
c_0 z + \frac{1}{2}q_0 z^2 + 
\sum_{i=1}^k (p_\theta)_i \left( c_b (z_b)_i + \frac{1}{2}q_b 
(z_b)_i^2
+ c_h (z_h)_i + \frac{1}{2}q_h (z_h)_i^2
\right ) \\
\subjectto \;\; & d - z\boldsymbol{1} \leq z_b,  \;\; z\boldsymbol{1} - d \leq z_h, \;\;
z,z_h, z_b \geq 0.
\end{split}
\end{equation}

To demonstrate the explicit formula for argmin operation
Jacobians for this particular case (e.g., to compute the terms in 
\eqref{eq-jacobian-equation-app}), note that we can write the above QP in inequality
form as $\minimize_{\{\boldsymbol{z}:G\boldsymbol{z} \leq h\}} \frac{1}{2} \boldsymbol{z}^T Q \boldsymbol{z} +
c^T \boldsymbol{z} $ with
\begin{small}
	\begin{equation}
	\boldsymbol{z} = \left [ \begin{array}{c} z \\ z_b \\ z_h \end{array} \right ], \; 
	Q = \left[\begin{array}{ccc} q_0 & 0 & 0 \\ 0 & q_b p_\theta & 0 \\ 0 & 0 & q_h
	p_\theta
	\end{array}
	\right ], \;
	c = \left [ \begin{array}{c} c_0 \\ c_b p_\theta \\ c_h p_\theta \end{array} \right ], \;
	G = \left[\begin{array}{ccc} -1 & -I & 0 \\
	1 & 0 & -I \\
	-1 & 0 & 0 \\
	0 & -I & 0 \\
	0 & 0 & -I \\
	\end{array} \right ],\; h = \left[ \begin{array}{c} -d \\ d \\0 \\0 \\ 0 
	\end{array} \right ].
	\end{equation}
\end{small}%
Thus, for an optimal primal-dual solution $(\boldsymbol{z}^\star, \lambda^\star)$, we
can compute the Jacobian $\frac{\partial \boldsymbol{z}^\star}{\partial p_\theta}$ (the
Jacobian of the optimal solution with respect to the probability vector
$p_\theta$ mentioned above), via the formula
	\begin{equation}
	\left [ \begin{array}{c} \frac{\partial \boldsymbol{z}^\star}{\partial p_\theta} \\ 
	\frac{\partial \lambda^\star}{\partial p_\theta} \end{array} \right ] = 
	\left [ \begin{array}{cc}
	Q & G^T \\
	D(\lambda^\star)G & D(G \boldsymbol{z}^\star - h) 
	\end{array} \right]^{-1} \left [ \begin{array}{c} 0 \\ q_b z_b^\star + c_b 1 \\
	q_h z_h^\star + c_h 1 \\ 0 \end{array} \right],
	\end{equation}
where $D(\cdot)$ denotes a diagonal matrix for an input vector.
After solving the problem and computing these Jacobians, we can compute the
overall gradient with respect to the task loss $L(\theta)$ via the chain rule
\begin{equation}
\frac{\partial L}{\partial \theta} = 
\frac{\partial L}{\partial \boldsymbol{z}^\star}
\frac{\partial \boldsymbol{z}^\star}{\partial p_\theta}
\frac{\partial p_\theta}{\partial \theta}
\end{equation}
where $\frac{\partial p_\theta}{\partial \theta}$ denotes the Jacobian of the
model probabilities with respect to its parameters, which are computed in the
typical manner.  
Note that in practice, these Jacobians need not be computed
explicitly, but can be computed efficiently via backpropagation; 
we use a recently-developed differentiable batch QP solver \cite{amos2017optnet}
to both solve the optimization problem in QP form and compute its derivatives.

\subsection{Details on computation for power scheduling problem}
The objective for the load forecasting problem is given by
\begin{equation}
\begin{split}
\minimize_{z \in \mathbb{R}^{24}} \;\;& \sum_{i=1}^{24} \mathbf{E}_{y \sim p(y | x;\theta)}\left [ \gamma_s [y_i-z_i]_+ + \gamma_e
[z_i-y_i]_+ + \frac{1}{2}(z_i - y_i)^2 \right ]\\
\subjectto \;\;& |z_i - z_{i-1}| \leq c_r \; \forall i,
\end{split}
\label{eq-gen-optim-appendix}
\end{equation}
where $z$ is the generator schedule, $y$ is the stochastic demand (which is affected by features $x$), $[v]_+ \equiv \max\{v, 0\}$, $\gamma_e$ is an over-generation penalty, $\gamma_s$ is an under-generation penalty, and $c_r$ is a ramping constraint. Assuming that $y_i$ is a Gaussian random variable with mean $\mu_i$
and variance $\sigma_i^2$, then this expectation has a closed form that can be
computed via analytically integrating the Gaussian PDF. 
Specifically, this closed form is
\begin{equation}
\begin{split}
\mathbf{E}_{y \sim p(y | x;\theta)}&\left [ \gamma_s [y_i-z_i]_+ + \gamma_e
[z_i-y_i]_+ + \frac{1}{2}(z_i - y_i)^2 \right ] \\
&= \underbrace{(\gamma_s + \gamma_e) (\sigma^2 p(z_i;\mu,\sigma^2) + (z_i-\mu) F
(z_i;\mu,\sigma^2))
- \gamma_s (z_i - \mu)}_{\alpha(z_i)} + \frac{1}{2}((z_i - \mu_i)^2 + \sigma_i^2),
\end{split}
\end{equation}
where $p(z;\mu,\sigma^2)$ and $F(z;\mu,\sigma^2)$ denote the Gaussian PDF and
CDF, respectively with the given mean and variance.  This is a convex function of $z$ 
(not apparent in this form, but readily established because it is an expectation
of a convex function), and we can thus optimize it efficiently and compute the
necessary Jacobians. 

Specifically, we use sequential quadratic programming (SQP) to iteratively
approximate the resultant convex objective as a quadratic objective, 
and iterate until convergence; specifically, we repeatedly solve
\begin{equation}
\begin{split}
z^{(k+1)} = \argmin_{z}\;\; & \frac{1}{2} z^T \mathrm{diag}\left(\frac{\partial^2 \alpha(z_i^{(k)})}{\partial z^2}  + 1\right) z + \left(\frac{\partial \alpha(z^{(k)})}{\partial z} - \mu\right)^T z \\
\subjectto \;\;& |z_i - z_{i-1}| \leq c_r \; \forall i
\end{split}
\label{eq:sqp}
\end{equation}
until $||z^{(k+1)} - z^{(k)}|| < \delta$ for a small $\delta$, where
\begin{equation}
\begin{split}
\frac{\partial \alpha}{\partial z} &= (\gamma_s + \gamma_e) F(z; \mu, \sigma) - \gamma_s,\\
\frac{\partial^2 \alpha}{\partial z^2} &= (\gamma_s + \gamma_e) p(z; \mu, \sigma).
\end{split}
\end{equation}

We then compute the necessary Jacobians using the quadratic
approximation~\eqref{eq:sqp} at the solution, 
which gives the correct Hessian and gradient
terms. 
We can furthermore differentiate the gradient and Hessian with respect
to the underlying model parameters $\mu$ and $\sigma^2$,
again using a recently-developed batch QP solver \cite{amos2017optnet}.

\subsection{Details on computation for battery storage problem}
The objective for the battery storage problem is given by
\begin{equation}
\begin{split}
\minimize_{z_{\text{in}}, z_{\text{out}}, z_{\text{state}} \in \mathbb{R}^{24}} \;\;& \mathbf{E}_{y \sim p(y | x;\theta)}\left [ \sum_{i=1}^{24} y_i (z_{\text{in}} - z_{\text{out}})_i
+ \lambda \left\lVert z_{\text{state}} - \frac{B}{2} \right\rVert^2
+ \epsilon \lVert z_{\text{in}} \rVert^2 + \epsilon \lVert z_{\text{out}} \rVert^2
\right ]\\
\subjectto \;\;& z_{\text{state}, i+1} = z_{\text{state}, i} -  z_{\text{out}, i} + \gamma_{\text{eff}}z_{\text{in}, i} \; \forall i, \;\; z_{\text{state}, 1} = B/2, \\
& 0 \leq z_{\text{in}} \leq c_{\text{in}}, \;\; 0 \leq z_{\text{out}} \leq c_{\text{out}}, \;\; 0 \leq z_{\text{state}} \leq B,
\end{split}
\label{eq-storage-optim-appendix}
\end{equation}
where $z_{\text{in}}, z_{\text{out}}, z_{\text{state}}$ are decisions over the charge amount, discharge amount, and resultant state of the battery, respectively; $y$ is the stochastic electricity price (which is affected by features $x$); $B$ is the battery capacity; $\gamma_{\text{eff}}$ is the battery charging efficiency; $c_{\text{in}}$ and $c_{\text{out}}$ are maximum hourly charge and discharge amounts, respectively; and $\lambda$ and $\epsilon$ are hyperparameters related to flexibility and battery health, respectively.

Assuming $y_i$ is a random variable with mean $\mu_i$, the expectation in the objective has a closed form:
\begin{equation}
\begin{split}
\mathbf{E}_{y \sim p(y | x;\theta)}&\left [ \sum_{i=1}^{24} y_i (z_{\text{in}} - z_{\text{out}})_i
+ \lambda \left\lVert z_{\text{state}} - \frac{B}{2} \right\rVert^2
+ \epsilon \lVert z_{\text{in}} \rVert^2 + \epsilon \lVert z_{\text{out}} \rVert^2
\right ] \\
&=  \sum_{i=1}^{24} \mu_i (z_{\text{in}} - z_{\text{out}})_i
+ \lambda \left\lVert z_{\text{state}} - \frac{B}{2} \right\rVert^2
+ \epsilon \lVert z_{\text{in}} \rVert^2 + \epsilon \lVert z_{\text{out}} \rVert^2.
\label{eq-storage-optim-expectation}
\end{split}
\end{equation}

We can then write this expression in QP form as $\minimize_{\{\boldsymbol{z}:G\boldsymbol{z} \leq h, \; A\boldsymbol{z} = b\}} \frac{1}{2} \boldsymbol{z}^T Q \boldsymbol{z} + c^T \boldsymbol{z} $ with
\begin{small}
	\begin{equation}
	\begin{split}
	&\boldsymbol{z} = \left [ \begin{array}{c} z_{\text{in}} \\ z_{\text{out}}\\ z_{\text{state}} \end{array} \right ], \; 
	Q = \left[\begin{array}{ccc} \epsilon I & 0 & 0 \\ 0 & \epsilon I & 0 \\ 0 & 0 & \lambda I
	\end{array}
	\right ], \;
	c = \left [ \begin{array}{c} \mu \\ -\mu \\ -\lambda B \boldsymbol{1} \end{array} \right ], \;\\
	&G = \left[\begin{array}{ccc} I & 0 & 0 \\
	-I & 0 & 0 \\
	0 & I & 0 \\
	0 & -I & 0 \\
	0 & 0 & I \\
	0 & 0 & -I \\
	\end{array} \right ],\; h = \left[ \begin{array}{c} c_{\text{in}} \\ 0 \\ c_{\text{out}} \\0 \\ B \\ 0
	\end{array} \right ], \;
	A = \left[\begin{array}{ccc} 0 & 0 & 0, \ldots, 0, 1 \\
	\gamma_{\text{eff}}D_1^T & -D_1^T & D_1^T-D_2^T \\
	\end{array} \right ] \;
	b = \left[ \begin{array}{c} B/2 \\ 0
	\end{array} \right ],
	\end{split}
	\end{equation}
	\label{eq:storage-qp}
\end{small}
where $D_1 = \left[ \begin{array}{c} I \\ 0 \end{array} \right ] \in \mathbb{R}^{24 \times 23}$ and 
$D_2 = \left[ \begin{array}{c} 0 \\ I \end{array} \right ] \in \mathbb{R}^{24 \times 23}$.

For this experiment, we assume that $y_i$ is a lognormal random variable (with mean $\mu_i$); thus, to obtain our predictions, we predict the mean of $\log(y)$ (i.e., we predict $\log(\mu)$).
After obtaining these predictions, we solve \eqref{eq:storage-qp}, compute the necessary Jacobians at the solution, and update the underlying model parameter $\mu$ via backpropagation, again using \cite{amos2017optnet}.

\subsection{Implementation notes}
For all linear models, we use a one-layer linear neural network 
with the appropriate input and output layer dimensions.
For all nonlinear models, we use a two-hidden-layer neural network,
where each ``layer'' is actually a combination of linear, batch norm \cite{ioffe2015batch}, 
ReLU, and dropout ($p=0.2$) layers with dimension 200.
In both cases, we add an 
additional softmax layer in cases where probability distributions 
are being predicted.

All models are implemented using PyTorch\footnote{\url{https://pytorch.org}} and employ the Adam optimizer \cite{kingma2014adam}.
All QPs are solved using a recently-developed
differentiable batch QP solver \cite{amos2017optnet},
and Jacobians are also computed automatically using backpropagation
via the same.

Source code for all experiments is available at \url{https://github.com/locuslab/e2e-model-learning}.

\end{document}